\documentclass[10pt,conference]{IEEEtran}

\usepackage[T1]{fontenc}
\usepackage[utf8]{inputenc}

\usepackage[hidelinks]{hyperref}
\usepackage{amsmath,amssymb,amsfonts}
\usepackage{graphicx}
\usepackage{booktabs}
\usepackage{siunitx}
\usepackage{bm,bbm}
\usepackage{algorithm2e}
\usepackage{subcaption}
\usepackage{cite}
\usepackage{url}
\usepackage{array}

\graphicspath{{Figures/PDF/}{Figures/PNG/}}

\title{Sequential Feature Selection for Efficient Landslide Segmentation from Multi-Spectral Data}

\author{
\IEEEauthorblockN{Arsalaan Ahmad, Oktay Karaku\c{s}, Paul L. Rosin}
\IEEEauthorblockA{
School of Computer Science and Informatics\\
Cardiff University, Cardiff, United Kingdom\\
Email: arslaan1970@gmail.com
}
}

\begin{document}

\maketitle

\begin{abstract}
Landslide detection from satellite imagery has advanced through deep learning, yet most models rely on large, highly correlated spectral-topographic inputs whose contributions remain poorly understood. The question of which channels are actually necessary has received surprisingly little attention. This matters: redundant or correlated inputs obscure physical interpretability, inflate computational overhead, and can actively degrade model performance through the Hughes Phenomenon. We present a systematic, explainable channel-selection framework for the Landslide4Sense benchmark, combining Sentinel-2 multispectral and ALOS PALSAR terrain data with 16 engineered spectral and structural indices. Rather than relying on conventional single-band drop tests, which evaluate channels in isolation and miss interaction effects, we apply Sequential Forward Floating Selection (SFFS) to iteratively build and prune a candidate feature pool using a lightweight U-Net++ proxy model. Beyond identifying a compact 8-channel subset that matches or exceeds the segmentation F1 of configurations using up to 30 channels, we use the selection process itself to interrogate which spectral and topographic features landslide models genuinely rely on, and what this reveals about the physical cues driving their predictions. We argue that SFFS represents a principled feature selection approach to input design in Earth observation, in contrast to the prevailing practice of appending every available band and hoping the model learns what to ignore.
\end{abstract}


\section{Introduction}

Landslides continue to cause significant damage in mountainous regions, and accurate records of past events remain essential for risk assessment and long-term monitoring. The Sentinel-2 mission \cite{DRUSCH201225} has played a central role in this effort, providing freely accessible multispectral imagery at resolutions suitable for detecting landslide scars across diverse terrain. Deep learning models, particularly U-Net-style architectures and their successors such as U-Net++, have become the dominant approach for processing this data. However, their increasing complexity makes it difficult to determine which physical signals are actually driving model predictions \cite{10.1007/978-3-319-24574-4_28}.

As these models have matured, a common pattern has emerged in how input feature sets are constructed. The 14 native Sentinel-2/ALOS PALSAR bands are typically used as the base, with engineered spectral indices (e.g., NDVI, NDMI, BSI) and structural features such as Grey and Canny edge filters appended incrementally. These additions are often justified by their physical relevance to landslide detection, including vegetation loss, moisture variation, and bare soil exposure \cite{montero2023standardized}.
Work on earthquake-triggered landslides in arid environments has also shown that red-band-derived optical changes can support rapid landslide mapping where vegetation-based contrast is limited \cite{massa2026toward}. Recent architectures such as RMAU-NET \cite{pham2025rmaunetresidualmultiheadattentionunetarchitecture} and related approaches \cite{le2023landslidedetectionsegmentationusing} extend this strategy to more than 20 input channels, typically reporting marginal performance improvements over the 14-band Sentinel-2/ALOS PALSAR baseline. However, performance gains tend to plateau or degrade as additional channels are introduced, suggesting diminishing returns from increasingly complex input configurations. The underlying assumption in this practice is that pre-computed spectral indices provide information that cannot be effectively learned from the raw bands alone.

This assumption warrants closer examination in the context of remote sensing. Common indices such as NDVI, NDMI, and BSI are deterministic transformations of the original Sentinel-2 bands and do not introduce new spectral information beyond a re-expression of existing signals \cite{montero2023standardized}. In addition, both raw bands and derived indices often exhibit high mutual correlation, particularly across Red Edge and vegetation-sensitive wavelengths. This creates a setting in which redundant features may obscure rather than enhance the underlying signal. Such conditions align with the Hughes phenomenon \cite{1054102}, where increasing input dimensionality relative to the available training data can degrade model performance \cite{pal2010feature}. Importantly, this effect is not captured by single-band or linear ablation studies, which evaluate features in isolation and fail to account for interaction effects between channels \cite{guyon2003introduction}.

A related limitation of large input configurations is reduced interpretability. Even when high-dimensional models perform well, it remains unclear which inputs contribute meaningfully to predictions. A model operating on up to 30 channels of mixed spectral, topographic, and structural inputs is difficult to interrogate when errors occur, making it challenging to distinguish informative signals from noise. This lack of transparency is particularly problematic in operational settings such as disaster response, where model outputs must be both reliable and explainable. Identifying a compact and physically interpretable subset of input features is therefore not only a question of efficiency, but also of understanding the decision-making process of the model \cite{guyon2003introduction}.

In this study, we address this problem using the Landslide4Sense benchmark dataset \cite{Ghorbanzadeh_2022}, which contains globally distributed landslide events across multiple triggering mechanisms and geographic regions. Rather than incrementally expanding the input feature set, we apply Sequential Forward Floating Selection (SFFS) \cite{sffs}, a greedy forward–backward feature selection algorithm that evaluates features in combination rather than isolation. Starting from the 14 Sentinel-2 bands and a set of engineered spectral and structural features \cite{montero2023standardized}, we use a lightweight U-Net++ proxy model \cite{zhou2018unetnestedunetarchitecture} to identify a minimal subset of channels that preserves or improves segmentation performance. This allows us to examine which features are truly necessary for landslide detection and what this reveals about the physical signals learned by the model.

The remainder of this paper is organised as follows. Section 2 presents the dataset details, including data sources and preprocessing steps. Section 3 describes the proposed methodology and modelling framework. Section 4 provides the experimental analysis and results, evaluating the performance of the proposed approach. Finally, Section 5 discusses the findings and concludes the paper, outlining key insights and directions for future work.


\section{Dataset Details}
\subsection{Landslide4Sense Dataset}

The Landslide4Sense benchmark provides a globally distributed development set of 3,799 multispectral patches, capturing landslide occurrences from 2015 to 2021. Each sample is formatted as a 14-channel array of shape $128 \times 128 \times 14$. The spectral component (B1–B12) originates from the Sentinel-2 mission, with native spatial resolutions that vary by band: 10\,m (B2–B4, B8), 20\,m (B5–B7, B11, B12), and 60\,m (B1, B9, B10). These are supplemented by ALOS PALSAR-derived slope (B13) and elevation (B14) layers \cite{Ghorbanzadeh_2022}.

A defining challenge of this dataset is extreme pixel-level sparsity. While approximately 58.4\% of patches contain landslide labels, event pixels constitute only $\approx$2.5\% of the total pixel count \cite{pham2025rmaunetresidualmultiheadattentionunetarchitecture}. Per-patch analysis reveals a long-tail distribution, where many samples contain less than 2\% positive pixels and some include as few as a single labelled pixel ($\approx$0.0061\% of the patch area). This imbalance suggests that standard segmentation models are prone to high false-positive rates, motivating the need for input features with strong class separability.

The dataset spans four geographically distinct regions: Iburi (Japan), Kodagu (India), Gorkha (Nepal), and Taitung (Taiwan), each characterised by different landcover types and triggering mechanisms, ranging from seismic activity to extreme rainfall. Landslide pixels generally exhibit higher reflectance than background areas, particularly in the Red (B4), Red Edge (B5), and Short-wave Infrared (B12) bands, due to vegetation removal and exposure of soil and bedrock. However, contrast is weaker in other bands, especially in the near-infrared region, where landslide and non-landslide surfaces can appear similar. This highlights the importance of combining spectral and topographic information, such as slope, to improve separability.

While a standardised catalogue of over 200 spectral indices exists \cite{montero2023standardized}, we restrict our study to a subset of engineered features to maintain computational tractability while demonstrating the effectiveness of the proposed framework.

Following the official benchmark protocol \cite{Ghorbanzadeh_2022}, training patches are drawn from the first spatial quarter of each study area, with the remaining three quarters reserved for evaluation. This spatial stratification ensures that training and test sets are geographically non-overlapping within each region, providing a conservative estimate of within-region generalisation. Cross-region generalisation, where models trained on one geographic area are evaluated on entirely distinct regions, remains an open challenge and is identified as a direction for future work.

\subsection{Engineered Channels}

\begin{table}[hbt]
    \centering
    \caption{Feature engineering: comprehensive list of 16 additional channels used in this study.}
    \label{tab:engineered}
    \begin{tabular}{l l}
        \toprule
        \textbf{New band(s)} & \textbf{Formula / Method} \\
        \cmidrule(lr){1-1} \cmidrule(lr){2-2}

        Bands 15--17$^*$ & $(x - x_{\min})/(x_{\max} - x_{\min})$ \\

        Band 18: NDVI & $(B8 - B4)/(B8 + B4)$ \\

        Band 19: NDMI & $(B8 - B11)/(B8 + B11)$ \\

        Band 20: NBR & $(B8 - B12)/(B8 + B12)$ \\

        Band 21: Gray & $(B2 + B3 + B4)/3$ \\

        Bands 22--23 & Gaussian / Median filters ($k=3$) \\

        Bands 24--25 & Image gradients (Sobel $G_x$, $G_y$) \\

        Band 26 & Canny edge detector \\

        Band 27: SAVI & $((B8 - B4) / (B8 + B4 + 0.5)) \times 1.5$ \\

        Band 28: EVI & $2.5 \times ((B8 - B4) / (B8 + 6 \times B4 - 7.5 \times B2 + 1))$ \\

        Band 29: NDWI & $(B3 - B8)/(B3 + B8)$ \\

        Band 30: MNDWI & $((B11 + B4) - (B8 + B2)) / ((B11 + B4) + (B8 + B2))$ \\

        \bottomrule
    \end{tabular}
    \\\vspace{0.2cm}$^*$ \textit{$x$ values correspond to B2, B3, and B4 respectively.}
\end{table}

We augment the original 14-channel input with 16 engineered channels that encode specific physical hypotheses related to landslide detection. These include vegetation-related indices (NDVI, EVI, SAVI), moisture-sensitive indices (NDMI), and soil exposure indicators (NBR) \cite{montero2023standardized}. Additional structural features, such as image gradients and edge detectors, are included to capture geometric patterns associated with landslide boundaries.

Although a much larger number of spectral indices has been proposed in the remote sensing literature, we do not aim to exhaustively evaluate the full catalogue. Instead, we select a compact subset of commonly used indices that map onto physically plausible landslide cues, including vegetation disturbance, moisture variation, and exposed material. This choice keeps the candidate pool computationally tractable for SFFS, avoids overwhelming the search with many highly correlated transformations of the same raw bands, and allows the analysis to focus on which engineered features add value beyond the original multispectral inputs.

A correlation analysis between the engineered and raw channels is presented later in the experimental section to examine redundancy patterns within the candidate pool.

By incorporating both raw bands and their derived transformations, we allow the selection framework to determine whether pre-computed indices provide complementary information or whether they are redundant with respect to the original spectral signals.

\begin{itemize}
    \item \textbf{Bands 15--17 (Normalization):} Scaled reflectance for the visible bands (B2--B4), intended to reduce sensitivity to illumination variability and topographic shadowing \cite{Ghorbanzadeh_2022}.
    
    \item \textbf{Bands 18--20 (Core Spectral Indices):} Standard indices capturing vegetation and moisture-related contrast (NDVI, NDMI, NBR) \cite{montero2023standardized}.
    
    \item \textbf{Band 21 (Intensity):} A grayscale composite derived from the visible bands to capture overall brightness contrast \cite{le2023landslidedetectionsegmentationusing}.
    
    \item \textbf{Bands 22--26 (Structural Features):} Smoothed representations, gradients, and edge maps derived from the grayscale image, intended to capture spatial structure and local texture \cite{pham2025rmaunetresidualmultiheadattentionunetarchitecture, le2023landslidedetectionsegmentationusing}.
    
    \item \textbf{Bands 27--30 (Additional Spectral Indices):} Further engineered indices capturing vegetation and water-related contrast (SAVI, EVI, NDWI, MNDWI) \cite{montero2023standardized}.
\end{itemize}

All engineered features are appended to the raw channels and treated as individual candidates during the channel-selection process.

\section{Methodology}
This section outlines the overall experimental design used to evaluate channel selection for landslide segmentation. We first establish a strong architectural and training baseline to ensure fair comparison across input configurations. Building on this foundation, we then introduce the channel-selection framework, which combines search-based and attribution-based analyses to identify a compact and effective subset of input bands.

\subsection{Architecture and Performance Baselines}

\subsubsection{\textbf{Benchmark Model}}

To establish a foundation for channel selection, we benchmark three configurations on the Landslide4Sense dataset: the official competition baseline (57.81\% F1), a standard U-Net (67.2\%), and U-Net++ with a ResNet-50 backbone \cite{he2016deep} (69.2\%). Given its superior performance, U-Net++ is adopted as the proxy model for all subsequent channel-selection experiments.

Although U-Net++ is used as the primary proxy model for SFFS, DeepLabV3+\cite{chen2017deeplab} with a ResNet-50 backbone is also evaluated as a secondary architecture to assess whether the selected channel subset remains competitive outside the model used during feature selection.

\subsubsection{\textbf{Training Strategy}}

As noted above, landslide detection is characterised by extreme class imbalance, with positive pixels accounting for only approximately 2.5\% of the dataset. To mitigate this, we employ a three-part training strategy, together with a fixed optimisation protocol across all channel-selection experiments. All models were trained using the Adam optimiser \cite{kingma2017adammethodstochasticoptimization}
under a fixed evaluation protocol with default PyTorch parameters. The official Landslide4Sense benchmark test partition was withheld entirely during feature selection and hyperparameter tuning. For all intermediate experiments, including SFFS subset evaluation and threshold optimisation, a fixed 80/20 train--validation split was generated from the training partition using random seed 42 and reused consistently across all runs.

\begin{itemize}
    \item \textbf{Loss Function:} While several prior studies have used focal loss to address class imbalance \cite{lin2018focallossdenseobject}, we instead use a combination of class-weighted Binary Cross-Entropy (BCE) and Dice loss \cite{sudre2017generalised}. This provides a balance between stable optimisation and overlap-sensitive segmentation performance.
    
    \item \textbf{Sampling:} We apply class-balanced sampling to increase exposure to positive tiles during training. In addition, occasional hard-negative mining is used to replace a portion of easy negative samples with more challenging background examples, improving the model’s ability to distinguish landslides from visually similar non-landslide regions.
    
    \item \textbf{Threshold Optimisation:} Because landslide segmentation is highly imbalanced, a fixed decision threshold of 0.5 is not necessarily optimal. Threshold selection was therefore performed exclusively on the validation split using a sweep over probability thresholds in the range 0.3--0.8. The threshold yielding the highest validation F1 score was retained for each experiment and subsequently applied during final evaluation.
\end{itemize}

During each SFFS evaluation round, candidate subsets are trained for 20 epochs under this same optimisation protocol, ensuring that performance differences are attributable to input composition rather than changes in training procedure.

Having defined a consistent training and evaluation protocol, we next examine how extending the input space with engineered features affects model performance. This provides a controlled reference point for subsequent channel-selection experiments.

\subsubsection{\textbf{Effect of Engineered Bands}}

To assess the contribution of engineered features, we extend the 14 raw channels with the additional engineered inputs described in Table~\ref{tab:engineered}. This expanded configuration serves as the reference model for subsequent channel-selection experiments, allowing any performance changes observed during SFFS to be attributed to input composition rather than architectural differences.

With a stable baseline and extended input configuration in place, the next step is to systematically reduce input dimensionality while preserving performance. To achieve this, we introduce a structured channel-selection framework that combines complementary analytical approaches.

\subsection{Channel-Selection Framework \& Implementation}

The objective is to identify a compact and task-oriented subset of input channels that preserves the performance of the full model while reducing redundancy and improving interpretability. To this end, we combine three complementary analyses: (i) Sequential Forward Floating Selection (SFFS) \cite{sffs}, (ii) channel redundancy analysis, (iii) Permutation Importance Test.

We begin with a search-based approach that evaluates feature subsets directly through model performance, allowing interactions between channels to be captured.

\subsubsection{\textbf{Sequential Forward Floating Search}}

SFFS is used as the primary channel-selection method because it evaluates candidate bands in combination rather than isolation, making it well suited to multispectral and topographic inputs with substantial redundancy and interaction effects. Starting from an empty set \(S = \emptyset\), the algorithm iteratively adds the band that maximises validation performance:

\begin{equation}
b^{\ast} = \arg\max_{b \notin S} \; F1\big(S \cup \{b\}\big),
\end{equation}

where the selected band is the one that gives the largest improvement when added to the current subset. This is followed by conditional backward elimination to remove redundant features. Unlike simple forward selection, SFFS includes a floating backward step after each inclusion, allowing previously selected bands to be removed if they become redundant once a new feature is added. This is important in multispectral settings, where correlated bands may appear useful individually but not jointly.

This forward--backward design makes SFFS more flexible than standard greedy selection while remaining computationally feasible for a 30-channel candidate pool. Because many engineered channels are deterministic transformations of the raw Sentinel-2 bands, SFFS is well-suited to identifying a compact subset of complementary inputs rather than retaining redundant ones.

The search terminates when no candidate addition, followed by floating elimination, yields a meaningful improvement in validation F1. The final subset is therefore interpreted as a compact set of mutually complementary features for landslide segmentation. To reduce instability arising from stochastic optimisation and severe class imbalance, the selection process was repeated across multiple runs using the same validation protocol, and the consistency of the selected subset was analysed across runs.

While SFFS identifies a performant subset, it does not explicitly quantify the relative importance of individual bands within that subset. To address this, we apply a complementary attribution-based analysis.

\subsubsection{\textbf{Channel Redundancy Analysis}}

To examine redundancy within the candidate feature pool, we computed a Pearson correlation matrix across the 30-channel input stack used in the feature-selection experiments. The analysis was computed using 1000 randomly sampled training patches, with 1024 pixels sampled per patch, yielding 1,024,000 pixel observations across 30 channels. Correlations were computed between channel values aggregated across the sampled pixels, providing a direct estimate of linear dependence between raw spectral bands, terrain variables, and engineered features. This analysis was used to support the interpretation of SFFS by identifying groups of channels that encode highly overlapping information.

\subsubsection{\textbf{Permutation Importance}}
To assess the relative contribution of each band within the final selected subset, we apply permutation-based feature importance estimation \cite{breiman2001random}. For each band, its values are randomly shuffled across the batch dimension, disrupting the spatial relationship between that channel and the target labels while preserving the overall value distribution. This approach is less susceptible to out-of-distribution artefacts than zero-masking, since the shuffled values remain within the observed data range.

To reduce shuffle variance, permutation is repeated ten times per band and the mean F1 score is recorded. The mean F1 drop relative to the unperturbed baseline is then computed across ten independent runs, and 95\% confidence intervals are derived using a t-distribution. A larger mean drop indicates a greater contribution of that band to segmentation performance within the selected subset.

Finally, we describe the implementation details that ensure the stability and reproducibility of the channel-selection process under the constraints of class imbalance and computational cost.

\subsubsection{\textbf{Implementation}}

All experiments were implemented in PyTorch \cite{paszke2019pytorch} and executed on an NVIDIA T4 GPU kernel in Google Colab. The U-Net++ model with a ResNet-50 backbone was used throughout. Models were trained using a batch size of 8 and an initial learning rate of $1 \times 10^{-4}$. No explicit weight decay or learning-rate scheduler was used. Input patches were resized from $128 \times 128$ to $256 \times 256$ prior to network input. Training augmentation consisted of random horizontal flips and random $90^\circ$ rotations, each applied with probability 0.5. Adam was used with default PyTorch parameters.

To stabilise SFFS under severe class imbalance, we employ a two-phase evaluation strategy:

\begin{itemize}
    \item \textbf{Pre-training:} The full input configuration is first trained to convergence in order to initialise the encoder with robust feature representations.
    
    \item \textbf{Iterative Selection:} During SFFS, the encoder is frozen, and only the decoder, together with a lightweight adapter layer, is trained for 20 epochs for each candidate subset.
    
    \item \textbf{Validation Stability:} The same validation split is used throughout selection, and experiments are repeated across multiple runs to assess the robustness of the selected subset.
    
    \item \textbf{Band Classification:} Bands are categorised as beneficial, detrimental, or redundant according to their effect on validation performance, using a tolerance threshold of $\pm 0.21\%$ F1 so that practically negligible differences are not over-interpreted during subset refinement.
    
\end{itemize}

\section{Experiments \& Results}

This section presents the experimental evaluation of the proposed channel-selection framework. We first analyse the subset of input bands identified by SFFS, then examine redundancy within the full candidate feature pool, followed by an attribution-based assessment using permutation importance. Finally, we compare the performance of the selected subset against several reference configurations to contextualise its effectiveness.

\subsection{SFFS-Selected Band Subset}

Applying SFFS yields a compact subset of eight input bands that preserves most of the segmentation performance achieved by the full model as seen in Table~\ref{tab:band_comparison}. The selected inputs consist of a small number of complementary spectral and terrain-derived features, while several redundant or unhelpful bands are consistently excluded during the floating backward steps.

Despite the substantial reduction in input dimensionality, from the full input configuration to an 8-channel subset, the reduced model achieves a slightly higher F1 score than the larger configurations. This suggests that accurate landslide segmentation does not require the full set of available raw and engineered inputs, and that principled subset selection can improve efficiency and interpretability without degrading performance.

Unlike single-band ablation or correlation-based pruning, SFFS explicitly evaluates bands in the context of a candidate subset. As a result, redundant features may be removed even when they appear individually informative, while bands that provide complementary information in combination are retained. The selected subset is analysed further through channel redundancy analysis and permutation importance in the following sections, before being compared against larger reference configurations.

\subsection{Channel Redundancy Analysis}

To assess whether the expanded input space contains redundant information, we computed a Pearson correlation matrix across the 30-channel candidate feature pool. The analysis used 1000 randomly sampled training patches, with 1024 pixels sampled per patch, yielding 1,024,000 pixel observations across 30 channels.

\begin{figure}[t]
    \centering
    \includegraphics[width=1\linewidth]{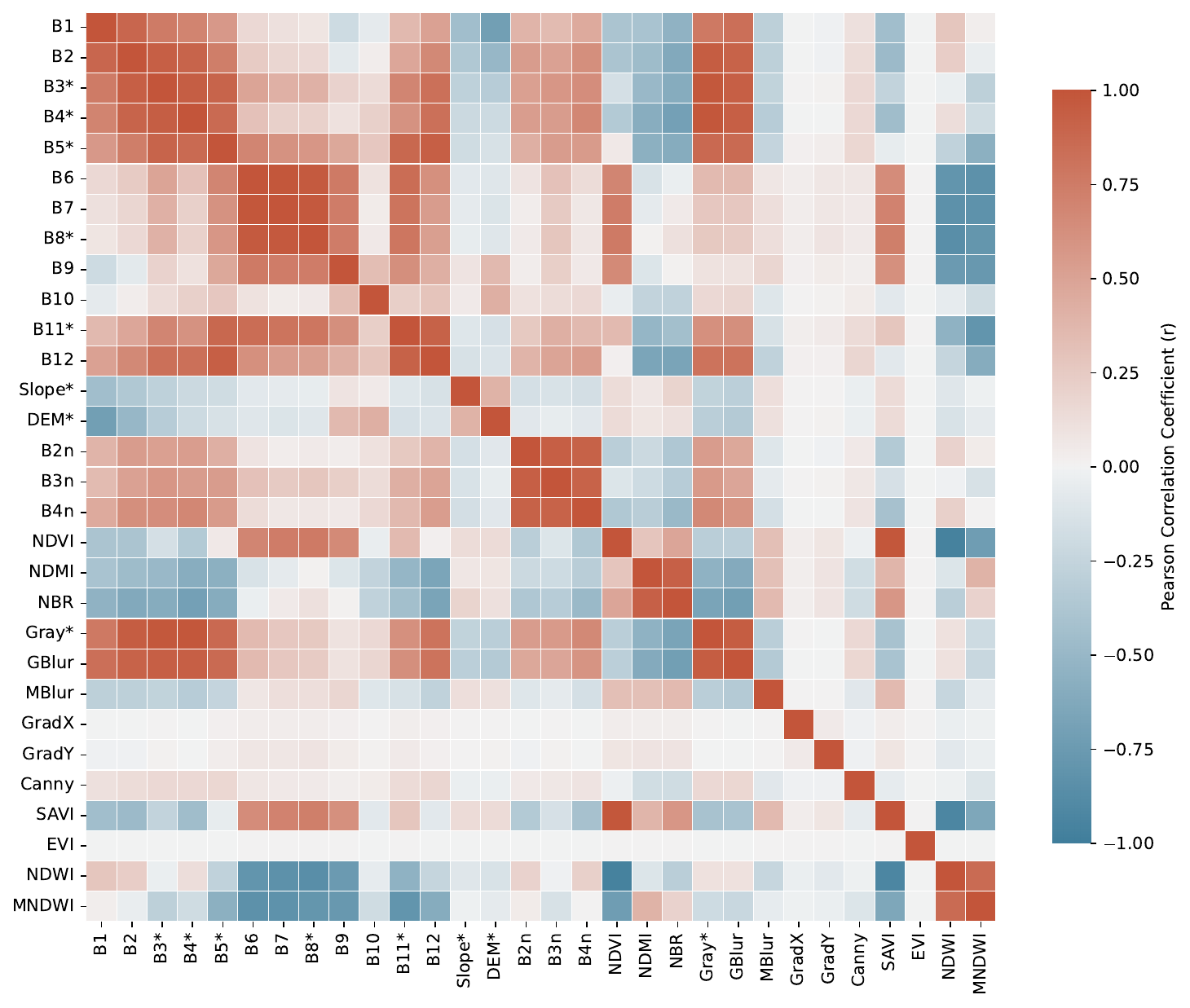}
    \caption{Pearson correlation matrix of the 30-channel candidate feature pool. The matrix was computed using 1000 training patches with 1024 sampled pixels per patch, yielding 1,024,000 pixel observations. Asterisks denote channels retained in the SFFS-selected subset. Strong correlation blocks are visible among adjacent spectral bands, visible-band transformations, grayscale/blurred representations, and engineered spectral indices, indicating substantial redundancy within the expanded feature set.}
    \label{fig:sffs}
\end{figure}

The correlation matrix reveals substantial redundancy within the expanded feature pool. Strong correlations are observed among adjacent red-edge/NIR bands, including B6--B7 ($r=0.988$) and B7--B8 ($r=0.974$), as well as between visible bands and grayscale-derived representations, such as B4--Gray ($r=0.987$), B3--Gray ($r=0.979$), and B2--Gray ($r=0.951$). Several engineered indices also show high interdependence, including NDVI--SAVI ($r=0.986$) and NBR--NDMI ($r=0.928$). These patterns support the view that many engineered channels are deterministic transformations of the original spectral bands and may not provide independent information. SFFS is therefore useful because it evaluates channels jointly, allowing correlated inputs to be excluded when they do not add complementary value to the selected subset.

\subsection{Permutation Importance}

To better understand the relative contribution of individual bands within the selected subset, we apply permutation importance analysis.

\begin{figure}[hbt]
    \centering
    \begin{minipage}{0.72\linewidth}
        \centering
        \includegraphics[width=\linewidth]{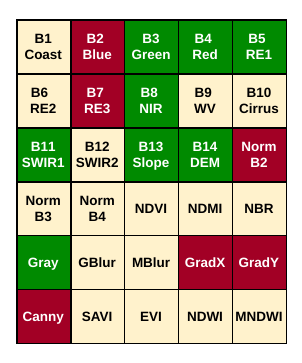}
        \caption{(a) SFFS-selected channel subset. Green cells indicate channels retained in the final compact configuration, while red cells denote channels consistently excluded during the floating backward steps.}
    \end{minipage}

    \vspace{0.6em}

    \begin{minipage}{0.72\linewidth}
        \centering
        \includegraphics[width=\linewidth]{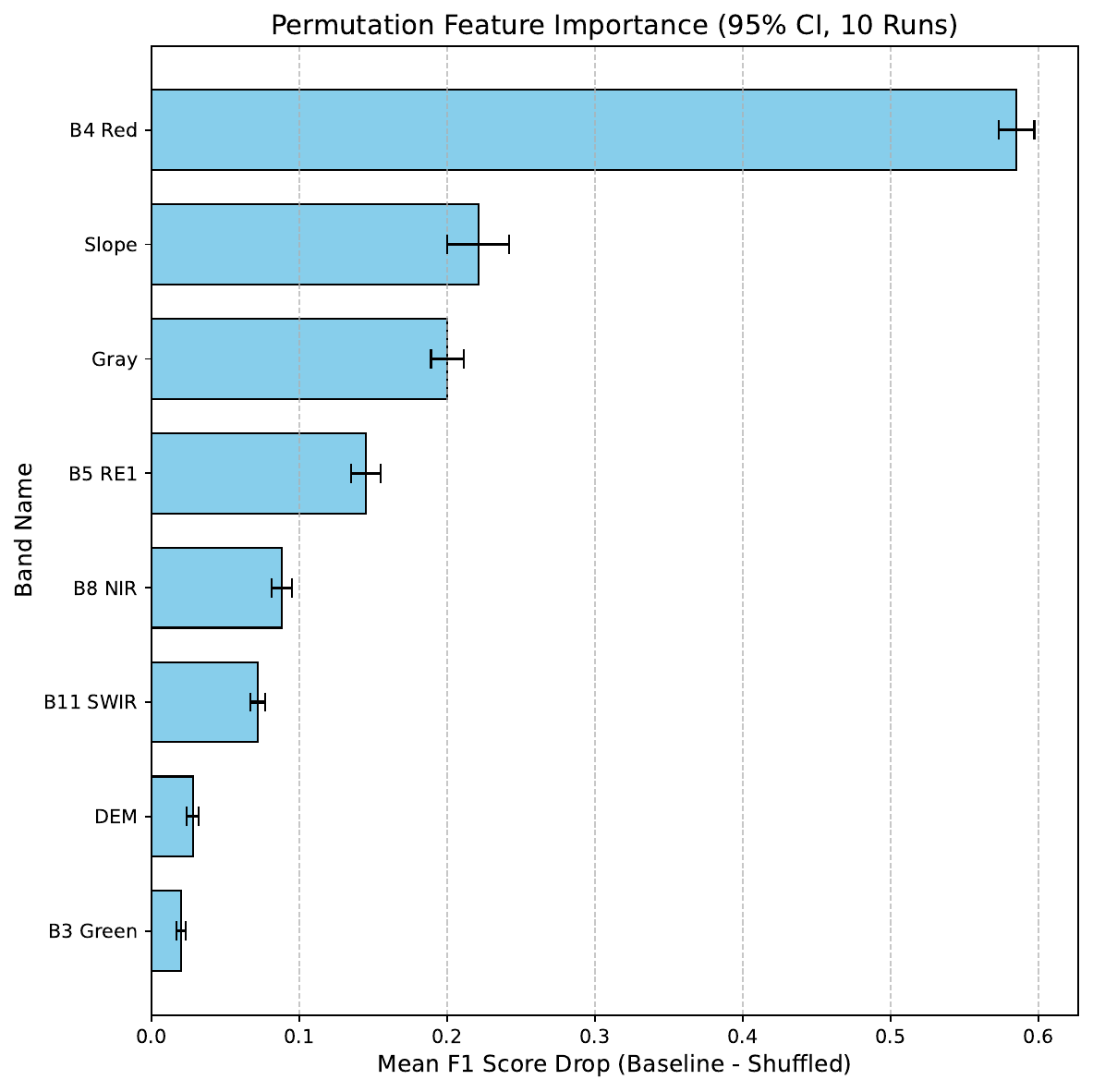}
        \caption{(b) Permutation feature importance across ten independent 
        runs. Bars show mean F1 drop when each band is permuted, with 
        95\% confidence intervals.}
    \end{minipage}

    \caption{Channel selection results. (a) SFFS-selected subset from 
    the full 30-channel candidate pool. (b) Permutation importance 
    of the selected 8-channel subset across ten independent runs.}
    \label{fig:selection_results}
\end{figure}

Permutation importance analysis shows that B4 (Red) is the dominant contributor within the selected subset, producing the largest mean drop in F1 when shuffled (Figure~\ref{fig:selection_results}b). Slope (B13) and the grayscale composite rank second and third, with substantially smaller but still notable effects. B5, B8, and B11 make moderate contributions, whereas DEM and B3 have only a limited impact on performance when permuted. The comparatively small contribution of DEM may reflect partial redundancy with slope rather than a complete lack of physical relevance. Overall, the confidence intervals remain narrow, particularly for the highest-ranked bands, indicating that the importance ordering is reasonably stable across runs rather than being driven by stochastic variation.
It is important to note that retention within the SFFS-selected subset does not imply equal conditional importance in the final model. A band may improve validation performance during subset construction, yet contribute only modestly under permutation once stronger complementary channels are already present.

\subsection{Final Results}

To contextualise the effectiveness of the selected subset, we compare it against several commonly used input configurations. 

\begin{table}[hbt]
\centering
\caption{Segmentation performance across channel configurations. Results are reported as mean F1 (\%) $\pm$ standard deviation across ten repeated runs.}
\label{tab:band_comparison}
\begin{tabular}{l c c c}
\toprule
\textbf{Configuration} & \textbf{\# Bands} & \textbf{U-Net++} & \textbf{DeepLabV3+} \\
\midrule
Raw spectral bands & 14 & $77.72 \pm 0.35$ & $74.90 \pm 0.41$ \\
Full engineered set & 30 & $77.30 \pm 0.46$ & $73.6 \pm 0.34$ \\
Referenced best set & 23 & $77.90 \pm 0.35$ & $75.1 \pm 0.23$ \\
\textbf{SFFS-selected subset} & \textbf{8} & $\mathbf{78.02 \pm 0.32}$ & $\mathbf{75.12 \pm 0.23}$ \\
\bottomrule
\end{tabular}
\end{table}

To contextualise the SFFS result, Table~\ref{tab:band_comparison} compares the selected subset with three reference configurations: the 14 raw spectral bands, the full 30-channel engineered set, and a 23-channel configuration comprising the 14 raw bands plus the first nine engineered channels, as used in recent Landslide4Sense studies \cite{pham2025rmaunetresidualmultiheadattentionunetarchitecture, le2023landslidedetectionsegmentationusing}. All reported results correspond to final evaluations on the official Landslide4Sense test partition after feature selection and threshold optimisation had been completed using only the training and validation data.

The SFFS-selected 8-channel subset achieves comparable or slightly better performance than the larger 23- and 30-channel configurations under U-Net++, while also remaining competitive under DeepLabV3+. This suggests that the selected subset is not solely tied to the U-Net++ architecture used during feature selection. The subset also provides clear computational benefits: using U-Net++ with a ResNet-50 encoder, it achieved 119.8 FPS and required 57.8 GFLOPs, compared with 80.4 FPS / 61.0 GFLOPs for the 14-channel baseline, 52.9 FPS / 64.2 GFLOPs for the 23-channel configuration, and 45.0 FPS / 71.0 GFLOPs for the full 30-channel feature set. All FPS and GFLOPs values were measured under the same inference setup.

Additionally, mean precision, mean recall, and mIoU followed the same overall pattern as F1 for the U-Net++ configurations. The selected 8-channel subset achieved comparable performance to the 14- and 23-channel settings (78.0\% precision, 74.0\% recall, and 61.2\% IoU), while providing the highest throughput at 119.8 FPS compared with 80.4 FPS and 52.9 FPS for the main U-Net++ configurations.

Overall, these results support the view that a compact input representation can offer a more efficient and interpretable alternative to larger feature stacks without compromising segmentation quality.

\section{Discussion \& Conclusion}

The main finding of this study is that an 8-channel input subset, selected through Sequential Forward Floating Search (SFFS), matches or slightly exceeds the segmentation performance of larger input configurations. Across ten independent U-Net++ runs, the selected subset achieved a mean F1 score of $78.02 \pm 0.32\%$, with individual runs ranging from 77.6\% to 78.4\%. This consistency suggests that the selection is stable rather than a product of stochastic training variation.

We interpret this performance pattern as consistent with the Hughes phenomenon \cite{1054102}, whereby increasing input dimensionality beyond a certain threshold leads to performance degradation due to redundancy and limited effective training signal. Notably, performance does not merely plateau but improves slightly in several runs when the input dimensionality is reduced, indicating that some channels in the full configuration may introduce noise rather than provide useful information.

\subsection{Physical Significance of the Selected Subset}

The SFFS results divide the input channels into three categories: a core subset of consistently selected features, a group of redundant features, and a set of channels whose inclusion reduces performance. The latter category is particularly notable, as these inputs are not simply ignored by the model but appear to negatively affect segmentation quality.

Among the retained channels, the selection is broadly consistent with known physical characteristics of landslides. Permutation importance further shows that B4 (Red) is the dominant contributor within the final subset, while slope (B13) is the strongest complementary terrain feature. This is physically plausible, as landslides often expose bare soil and damaged vegetation, increasing contrast in the red portion of the spectrum, while slope provides essential topographic context for where failures are most likely to occur. SWIR 1 (B11) and DEM (B14) also contribute terrain- and moisture-related information \cite{Ghorbanzadeh_2022}. The joint selection of B5 (Red Edge) and B8 (NIR) may indicate that these bands capture complementary aspects of vegetation disturbance \cite{delegido2011evaluation}, although this cannot be confirmed directly from segmentation performance alone. The inclusion of the grayscale composite (Band 21), despite being a linear combination of visible bands, suggests that simplified representations of brightness contrast may be more useful than multiple correlated inputs.
Importantly, retention within the SFFS-selected subset does not imply equal conditional importance in the final model: permutation analysis shows that some retained bands play a dominant role, whereas others contribute more modestly once stronger complementary cues are already present.

The behaviour of excluded channels is less straightforward to interpret. The exclusion of B2 (Blue) and its normalised counterpart may be related to atmospheric scattering effects in mountainous regions \cite{DRUSCH201225}, although this explanation remains tentative. Similarly, the exclusion of B7 (Red Edge 3) may reflect redundancy or interaction effects with other vegetation-sensitive bands, but a definitive physical interpretation is not clear.

Notably, only one of the engineered channels (the grayscale composite) is retained in the final subset. Most engineered spectral indices (NDVI, NDMI, NBR, SAVI, EVI, NDWI, MNDWI) are either redundant or negatively impact performance. This aligns with the observation that these indices are deterministic transformations of the original spectral bands and do not introduce new information. This interpretation is further supported by the redundancy analysis, which shows strong correlations among several engineered indices and their source spectral bands, including NDVI--SAVI and visible-band--grayscale relationships. While this finding should not be overgeneralised, it suggests that the common practice of appending spectral indices to deep learning inputs may not always be beneficial.

Structural features such as gradients and edge detectors are also excluded. This may indicate that the model is able to learn boundary information directly from spectral contrast, without requiring pre-computed structural cues. However, this observation is dataset- and architecture-dependent and requires further validation.

\subsection{Interpretability and Model Complexity}

Recent work in landslide detection has explored increasingly large input configurations and complex architectures \cite{pham2025rmaunetresidualmultiheadattentionunetarchitecture, le2023landslidedetectionsegmentationusing}. The results presented here suggest that such approaches may yield diminishing returns. Consistent with findings in broader remote sensing literature \cite{MA2019166, li2022dimensionality}, the relevance of input features appears to be more important than their quantity.

More broadly, the results suggest that channel selection should not be viewed solely as a dimensionality-reduction strategy. The selected subset can also be interpreted as a compact task-oriented representation of landslide-relevant information. Rather than retaining all available spectral and engineered features, SFFS identifies a set of complementary signals associated with vegetation disturbance, moisture variation, exposed material, terrain morphology, and intensity structure. In this sense, the objective is not simply to reduce the number of inputs, but to retain the information most relevant to the segmentation task while discarding redundant or uninformative channels.

This perspective aligns with recent trends in remote sensing, where performance improvements are increasingly attributed to the quality and complementarity of learned representations rather than the quantity of available inputs. Similar principles have been explored through spatial-frequency collaborative representations \cite{deng2025spatial}, prototype-based segmentation frameworks \cite{deng2026tri}, sparsity-aware learning strategies \cite{zhang2026tlsa,zhang2021balance}, and recent perception-driven remote sensing interpretation frameworks \cite{zhang2026glance}, which seek to capture task-relevant information while suppressing redundant or less informative signals.

The SFFS-selected subset achieves $78.02 \pm 0.32\%$ F1 using 8 channels, compared with $77.72 \pm 0.35\%$ for the 14-channel raw configuration and $77.90 \pm 0.35\%$ for the 23-channel reference configuration. While the absolute performance difference is modest, the reduction in input dimensionality is substantial, which may reduce storage and preprocessing requirements and improve computational efficiency as seen by the significant improvement in FPS and GFLOPs.

There are also clear interpretability benefits. A model operating on a small number of physically meaningful channels is easier to analyse and explain, particularly in operational settings such as disaster response. When predictions fail, it becomes possible to reason about which physical signals contributed to the decision, an advantage that is difficult to maintain in high-dimensional input spaces.

The identified subset may also be easier to transfer across related multispectral platforms, as the retained channels correspond to commonly available spectral and topographic cues rather than a highly specialised engineered stack. However, this study does not directly evaluate cross-sensor transfer, so this potential advantage should be interpreted cautiously.

Finally, reduced input dimensionality may be advantageous in resource-constrained settings. This observation is supported by the efficiency analysis presented in Section 4.4, where the selected subset achieved the highest inference throughput and lowest computational cost among all evaluated configurations. These results suggest that careful feature selection can improve both interpretability and deployment efficiency without sacrificing segmentation performance. Although runtime was not evaluated explicitly in this study, smaller input configurations are likely to be easier to deploy than larger feature stacks when memory or computational capacity is limited.

\subsection{Limitations and Future Work}

This study is conducted exclusively on the Landslide4Sense benchmark. Although the dataset is geographically diverse, spanning multiple climatic and tectonic regimes, further validation on additional datasets would strengthen the generalisability of the findings.

Future work may also investigate whether similar channel subsets are selected across different model architectures, which would help determine whether the observed patterns reflect underlying physical signals or architecture-specific behaviour. Additionally, extending the analysis to larger feature pools could provide further insight into the role of engineered indices in deep learning-based remote sensing.

\section*{Conflict of Interest Statement}

The authors declare that the research was conducted in the absence of any commercial or financial relationships that could be construed as a potential conflict of interest.

\section*{Author Contributions}

\textbf{AA}: Conceptualization, Data curation, Formal analysis, Investigation, Methodology, Software, Validation, Visualization, Writing – original draft.\\
\textbf{OK}: Conceptualization, Methodology, Supervision, Validation, Writing – review \& editing.\\
\textbf{PR}: Conceptualization, Methodology, Supervision, Validation, Writing – review \& editing, Project administration, Resources.

\section*{Funding}
This work was supported by a self-proposed research internship grant awarded to AA by Cardiff University under the supervision of PR.

\section*{Acknowledgments}
The authors would like to thank Cardiff University, School of Computer Science and Informatics for providing computational resources and supporting this research.

\small
\bibliographystyle{IEEEtran}
\bibliography{references}
\end{document}